\documentclass[pdflatex,sn-apa]{sn-jnl}% APA Reference Style

\usepackage{graphicx}%
\usepackage{multirow}%
\usepackage{amsmath,amssymb,amsfonts}%
\usepackage{amsthm}%
\usepackage{mathrsfs}%
\usepackage[title]{appendix}%
\usepackage{xcolor}%
\usepackage{textcomp}%
\usepackage{manyfoot}%
\usepackage{booktabs}%
\usepackage{listings}%
\usepackage[cjk]{kotex}
\usepackage{arydshln}
\usepackage{setspace}
\usepackage{algorithmic,algorithm}

\theoremstyle{thmstyleone}%
\theoremstyle{thmstyletwo}%
\theoremstyle{thmstylethree}%
\begin{document}

\title[Enriching KoLLA]{Enriching the Korean Learner Corpus with Multi-reference Annotations and Rubric-Based Scoring}

\author[1]{\fnm{Jayoung} \sur{Song}}\email{jayoung.song@psu.edu}
\author[2]{\fnm{KyungTae} \sur{Lim}}\email{ktlim@kaist.edu}
\equalcont{These authors contributed equally to this work.}
\author*[3]{\fnm{Jungyeul} \sur{Park}}\email{jungyeul@mail.ubc.ca}
\affil[1]{\orgdiv{Department of Asian Studies}, \orgname{The Pennsylvania State University}, \orgaddress{\street{201 Old Main}, \city{University Park}, \postcode{16802}, \state{PA}, \country{USA}}}
\affil[2]{\orgdiv{Graduate School of Culture Technology}, \orgname{Korea Advanced Institute of Science \& Technology}, \orgaddress{\street{291 Daehak-ro}, \city{Daejeon}, \postcode{34141}, \country{South Korea}}}
\affil[3]{\orgdiv{Department of Linguistics}, \orgname{The University of British Columbia}, \orgaddress{\street{2613 West Mall}, \city{Vancouver}, \postcode{V6T 1Z4}, \state{BC}, \country{Canada}}}

% \author{LRE anonymous submission}

\abstract{
Despite growing global interest in Korean language education, there remains a significant lack of learner corpora tailored to Korean L2 writing. To address this gap, we enhance the KoLLA Korean learner corpus by adding multiple grammatical error correction (GEC) references, thereby enabling more nuanced and flexible evaluation of GEC systems, and reflects the variability of human language. Additionally, we enrich the corpus with rubric-based scores aligned with guidelines from the Korean National Language Institute, capturing grammatical accuracy, coherence, and lexical diversity. These enhancements make KoLLA a robust and standardized resource for research in Korean L2 education, supporting advancements in language learning, assessment, and automated error correction.
% This paper presents an enriched version of the KoLLA Korean learner corpus, introducing two key enhancements: multi-reference GEC (GEC) annotations and rubric-based scoring aligned with the Korean National Language Institute's guidelines. The dual-reference setup addresses the linguistic variability inherent in Korean and provides a more flexible and robust evaluation framework for GEC systems. Meanwhile, rubric-based scoring moves beyond holistic judgments to structured evaluations of learner essays, improving the precision and interpretability of proficiency assessments. Together, these enhancements position KoLLA as a standardized and comprehensive resource for research in Korean L2 education and computational language learning.
}




\maketitle

% \tableofcontents

\section{Introduction}

Written and spoken data produced by second language (L2) learners have long served as important resources for studying L2 acquisition. With the development of electronic collections of learner data—known as learner corpora—researchers now benefit from more systematic and scalable analysis methods. These corpora, often large and representative, offer a stronger empirical foundation than smaller, manually compiled datasets. They improve the reliability of research by enabling large-scale analysis of learner errors, interlanguage patterns, and developmental stages of acquisition \citep{zhang-fu-2024-review}. In addition, advanced computational tools make it possible to perform quick and detailed analyses, facilitating broader investigations into L2 learning \citep{lebruyn-paquot-2021-learner}.

The increasing availability of learner corpora and reliable automated analytical tools has facilitated a broad range of empirical studies. These studies leverage corpus linguistic methods to investigate diverse aspects of L2 acquisition, such as error patterns, lexical diversity, and syntactic complexity. For instance, error editors and annotation tools allow researchers to insert detailed error information into text files, providing insights into the specific challenges learners face at various proficiency levels. Tools like the UCLouvain Error Editor and automatic annotation frameworks such as ERRANT have been instrumental in standardizing and automating the error annotation process across different languages \citep{bryant-etal-2017-automatic, bryant:2019}.

While learner corpus research has significantly advanced for widely taught languages like English and Chinese—particularly in the area of grammatical error correction (GEC)—Korean remains underrepresented in this domain. This gap is notable given the rapid global expansion of Korean language education, driven by the international popularity of Korean culture, including K-pop, dramas, and films. The growing number of Korean L2 learners highlights the urgent need for robust linguistic resources to support their language development. According to the Modern Language Association, Korean ranked among the top ten most-taught languages in U.S. higher education institutions between 2013 and 2016, underscoring its increasing global significance \citep{lim-song-park-2023}.\footnote{\url{https://www.mla.org/content/download/191324/file/Enrollments-in-Languages-Other-Than-English-in-US-Institutions-of-Higher-Education-Fall-2021.pdf?utm_source=chatgpt.com}}

Existing Korean learner corpora, such as KoLLA (Korean Learner Language Annotations) and the KyungHee Korean Learner Corpus, have made notable contributions to this field. KoLLA, developed through open writing assignments by Korean L2 learners, has been used to study grammatical errors and learner writing performance \citep{dickinson-israel-lee-2010-building,israel-dickinson-lee:2013:IJCNLP}. Despite these efforts, the scope and utility of these corpora remain limited due to the absence of multi-reference GEC (GEC) data and rubric-based scoring aligned with Korean language education standards.

In this paper, we present an enhanced version of the existing KoLLA corpus, originally developed by Indiana University and Wellesley College. Our work systematically augments KoLLA in two key ways. First, we introduce a layer of multi-reference grammatical error corrections (GEC), transforming the dataset into a multi-reference GEC corpus. This design aligns with best practices in GEC research, where acknowledging linguistic variability across corrections is critical for robust system evaluation. Second, we add rubric-based scores using standardized criteria issued by the Korean National Language Institute. These rubric scores offer a more nuanced and structured approach to writing assessment, supporting comparative evaluation and instructional design.

The enriched KoLLA corpus consists of 100 essays written by Korean L2 learners, including both heritage and non-heritage speakers, at beginner and intermediate levels. These essays yield 1,419 segmented sentences, with an average of approximately 14 sentences per learner and one essay per participant. Each sentence is accompanied by two reference corrections, constituting a total of 2,838 GEC references, making KoLLA the first publicly available Korean learner corpus with multi-reference GEC and standardized rubric scores.

This paper introduces these enhancements, clarifies the rationale behind multi-reference annotations and rubric standardization, and outlines the procedures followed to ensure annotation quality and inter-annotator reliability. We further clarify that this work constitutes a significantly enriched version of the original KoLLA dataset, and should be considered a new release that builds upon and extends the earlier resource.

\section{Previous Works} \label{previous-work}

GEC and Automated Essay Scoring (AES) have been extensively researched for widely taught languages such as English and Chinese, leading to notable advancements in learner corpus development, grammatical error annotation \citep{bryant-etal-2017-automatic,lin-2024-evaluating,gu-etal-2025-improving}, and the incorporation of holistic and rubric-based scoring frameworks \citep{ridley-EtAl:2021:AAAI,he-etal-2022-automated}.
These studies have contributed resources for understanding the linguistic challenges faced by L2 learners and enhancing tools for language education. However, research and resources for Korean GEC and AES remain scarce, with only a limited number of learner corpora and studies dedicated specifically to the Korean language.

In the context of GEC, the development of annotated learner corpora has played a critical role in enabling detailed analyses of learner errors and supporting the development of correction systems. For instance, large-scale learner corpora like the  First Certificate in English (FCE) corpus, a subset of the Cambridge Learner Corpus (CLC) for English \citep{yannakoudakis-briscoe-medlock:2011:ACL-HLT} and the Multi-Reference Multi-Source Evaluation Dataset (MuCGEC) for Chinese \citep{zhang-etal-2022-mucgec} have provided the foundation for building GEC systems that can address a wide range of grammatical issues. 
These corpora typically include  corrections with or without error annotations allowing us to analyze error patterns across proficiency levels and develop targeted instructional strategies. 

In the context of AES, the development of annotated datasets has been instrumental in facilitating the evaluation of writing proficiency and supporting the design of automated scoring systems. For example, the Automated Student Assessment Prize (ASAP) dataset has been a foundational resource in AES research. ASAP provides a collection of student essays scored according to rubric scores across several prompts. This dataset has enabled us to train and evaluate models that assess the overall quality of essays, taking into account factors such as coherence, fluency, and vocabulary. To further enrich this resource, the ASAP++ dataset was introduced, enhancing the original ASAP dataset with additional essay attribute scores, such as sentence structure, word choice, and organization \citep{mathias-bhattacharyya:2018:LREC}. These fine-grained rubric score annotations allow for a more detailed analysis of writing attributes and enable systems to provide more nuanced feedback to learners. Similar to learner corpora in GEC, datasets like ASAP and ASAP++ provide the foundation for empirical studies and practical applications in AES. These resources support the development of automated scoring systems capable of assessing diverse linguistic features and addressing the needs of both learners and educators.

For Korean, however, relatively few studies have focused on developing learner corpora for GEC \citep{yoon-etal-2023-towards} and AES \citep{lim-song-park-2023}. Among the notable exceptions are the KoLLA and Kyunghee Korean learner corpora, both of which have contributed to the field but remain limited in scope and scale. The KoLLA corpus, developed by Indiana University and Wellesley College, represents one of the earliest efforts to construct a Korean learner corpus with detailed error annotations. It comprises approximately 12,000 words and 1,419 sentences drawn from 100 essays, categorized by learner proficiency (beginner and intermediate) and learner type (heritage and non-heritage) \citep{lee-dickinson-israel-2012-developing}. KoLLA places particular emphasis on annotating errors in functional morphemes—such as postpositions and verbal endings—which are widely recognized as especially challenging for learners of Korean.
The Kyunghee Korean learner corpus \citep{park-lee:2016}, is a significant large-scale resource containing 3,700 writing samples from Korean language learners across various proficiency levels. It offers insights into linguistic features that characterize L2 learners’ proficiency levels. This corpus has been employed in tasks such as POS tagging, language modeling, and syntactic parsing, contributing to a deeper understanding of learner language. However, the corpus lacks essential developments in error annotation and multi-reference GEC data, which are crucial for advancing GEC and AES systems. Recent studies, such as \citet{lim-song-park-2023}, have explored its potential for scoring systems, but the corpus remains limited by offering only holistic scores instead of detailed rubric-based scores. This limitation restricts its utility in capturing broader linguistic variability and developing more robust evaluation frameworks.

This study aims to address these limitations by enhancing the KoLLA corpus to create a more comprehensive resource for Korean GEC and AES. Specifically, we augment the corpus with multi-reference GEC and rubric scores aligned with the guidelines of the Korean National Language Institute. These enhancements position the KoLLA corpus as a robust and standardized evaluation dataset for Korean L2, bridging the gap between current resources and the increasing demand for advanced Korean learner corpus research.

\section{Development and Annotation} \label{new-datasets}

This section introduces two key datasets that enhance the KoLLA learner corpus: a multi-reference GEC dataset and a rubric-based AES dataset. These additions provide insights into Korean L2 learning and assessment, addressing critical gaps in existing resources. The multi-reference GEC dataset incorporates multiple corrected versions for each sentence, capturing linguistic variability and enabling more comprehensive grammatical error analysis. The rubric-based AES dataset, aligned with the Korean National Language Institute guidelines, transitions from holistic scoring to detailed rubric-based evaluations, improving the reliability of L2 writing assessments. Together, these datasets establish the KoLLA corpus as a robust resource for advancing Korean L2 acquisition research, supporting innovations in language learning, teaching, and automated assessment.

\subsection{Introduction to KoLLA}

The KoLLA learner corpus addresses the challenge of annotating and analyzing Korean post-positional particle errors, a major difficulty for learners of Korean as a L2 \citep{lee-dickinson-israel-2012-developing}. These particles, akin to English prepositions, serve multiple functions, including marking grammatical roles (e.g., subject and object), semantic roles (e.g., agent and location), and discourse functions. The agglutinative nature of Korean and the tendency for particles to combine with nominals without whitespace complicate their correct use for learners.
The KoLLA corpus annotates essays written by college-level Korean learners, grouped by proficiency (beginner and intermediate) and learner type (heritage and non-heritage). The annotation scheme is multi-layered, addressing spacing and spelling errors to ensure proper segmentation of words before particle errors are classified. Errors are categorized into four types: omission, substitution, addition, and ordering.
To determine the correct particles for annotation, the corpus adopts the principle of “minimal interaction,” focusing on grammatical forms expected in formal Korean writing. Additionally, it accounts for cases where multiple particles may be equally valid, providing a set of possible correct forms for robust system evaluation.
Developed using 100 learner essays and tools like EXMARaLDA \citep{schmidt-2010-linguistic}, the corpus achieves high inter-annotator agreement rates, validating its guidelines and methodology. By supporting automatic particle error detection, KoLLA aims to advance the analysis of morphologically rich learner languages. The resource also contributes to broader computational linguistic efforts by offering insights into annotation best practices for error detection and language learning applications.

The dataset used in this study is based on the original KoLLA corpus, which consists of 100 essays written by Korean L2 learners. These essays were segmented into 1,419 individual sentences to align with the differing evaluation levels required by AES and GEC systems. While AES operates at the essay level, GEC is typically conducted at the sentence level, thus necessitating this segmentation.

To support multi-reference evaluation for GEC, each of the 1,419 sentences is accompanied by two human-generated corrections. The first correction is sourced from the original KoLLA corpus, while the second was newly created by a trained linguist following detailed annotation guidelines. This dual-reference design enhances the resource’s ability to capture the range of valid corrections—especially important in a morphologically rich language like Korean.
For AES, two annotators independently assigned rubric-based scores to all 100 essays, evaluating dimensions such as content, coherence, grammatical accuracy, and vocabulary use.

The present version constitutes a substantial enrichment of the original KoLLA corpus, rather than an entirely new resource. While the learner-produced texts are drawn from the original KoLLA dataset, this version introduces additional multi-reference GEC annotations and double-coded AES scoring, following best practices in learner corpus construction and evaluation. The enriched KoLLA resource thus supports more comprehensive investigations into both grammatical error correction and essay scoring for Korean as a foreign language.

\subsection{Multi-reference GEC corpus} \label{gec-corpus}

The KoLLA corpus, consisting of open writing assignments from Korean L2 learners, serves as the foundation for our enhanced multi-reference GEC dataset. The original corpus includes 1,419 sentences written by beginner and intermediate-level learners, segmented from essays to facilitate sentence-level GEC annotation, given the difficulty of automatic boundary detection in the presence of learner errors.

To create a multi-reference GEC resource, we augmented the original KoLLA corrections with a second human-generated correction for each sentence. This design follows precedents in English and Chinese GEC corpora (e.g., FCE, MuCGEC), where multiple references are employed to capture the inherent ambiguity and variability of learner errors. For Korean, such variability is even more pronounced due to its morphosyntactic flexibility, honorific system, and case-marking ambiguity. In agglutinative languages, a single correction may fail to represent the full range of valid alternatives, potentially penalizing correct but divergent model outputs.

The second layer of annotation was produced by a trained linguist holding an M.A. in Korean Language Education. This annotator followed detailed guidelines to ensure fluency, preserve the learner’s intended meaning, maintain structural fidelity, and avoid unnecessary rewriting. Although the second set of corrections was provided by a single annotator, quality assurance focused on internal consistency through strict adherence to annotation guidelines.

The resulting dataset thus provides two reference corrections per sentence, offering a more realistic and flexible evaluation framework for Korean GEC. Multi-reference annotation improves the corpus’s ability to capture correction variability and supports more realistic evaluation settings for Korean GEC.

Grammatical error typology provides a principled framework for systematically identifying, classifying, and analyzing learner errors, serving as a foundation for building and evaluating GEC systems. Building on the original MRU framework (Missing, Replacement, and Unnecessary) introduced by \texttt{errant} \citep{bryant-etal-2017-automatic}, we extend it for Korean by incorporating language-specific error types and annotation strategies, formalized in Algorithm~\ref{error-classification-algorithm}.

A key extension involves the treatment of word boundary errors, which are prevalent in agglutinative languages such as Korean. We introduce two fine-grained categories: \texttt{M:WB} (missing word boundaries) and \texttt{U:WB} (unnecessary word boundaries). \texttt{M:WB} is applied when multiple lexical items are erroneously fused into a single token (typically due to omitted spacing), while \texttt{U:WB} is used when excessive spacing disrupts a morphosyntactically valid token.

Spelling errors are labeled as \texttt{R:SPELL} when a content word is incorrectly realized orthographically or phonologically. We further expand the framework to explicitly address errors involving functional morphemes—especially postpositions (\texttt{ADP}) and verbal endings (\texttt{PART}). If the associated content word is correct, such errors are marked according to their context: a missing postposition is annotated as \texttt{R:POS$_i$~$\rightarrow$~POS$_j$+ADP}, while an unnecessary postposition is marked as \texttt{R:POS$_i$+ADP~$\rightarrow$~POS$_j$}. Substitution of one postposition for another is marked as \texttt{R:POS+ADP$_{i}$~$\rightarrow$~POS+ADP$_{j}$}. When the associated content word is itself misspelled or incorrect, the error is subsumed under \texttt{R:SPELL}, and the morpheme is not annotated independently.

To mitigate segmentation mismatches and annotation inconsistencies, preprocessing was applied to normalize common variants in particle use and compound verb boundaries. This step reduced word boundary misalignment errors to under 5\%.

Finally, word order errors are annotated as \texttt{R:ORDER} when the source and target sentences contain the same lexical items but differ in sequence, indicating syntactic reordering without lexical substitution.

\vspace{1em}
Algorithm~\ref{error-classification-algorithm} formalizes this annotation logic and accounts for the structural and orthographic complexities characteristic of Korean grammatical errors.

\begin{algorithm}[!ht]
\caption{Error annotation procedure} \label{error-classification-algorithm}
{\footnotesize
\begin{algorithmic}[1]
\STATE \textbf{Given a source and target sentence, perform the following checks:}

\IF{Associated content word is correct}
    \IF{ADP or PART is present unnecessarily in the source sentence}
        \STATE $\mathcal{A} =$ \texttt{R:POS$_i$+ADP~$\rightarrow$~POS$_j$}
    \ELSIF{ADP or PART is missing in the source sentence}
        \STATE $\mathcal{A} =$ \texttt{R:POS$_i$~$\rightarrow$~POS$_j$+ADP}
    \ELSIF{ADP or PART is incorrect (substitution needed)}
        \STATE $\mathcal{A} =$ \texttt{R:POS+ADP$_{i}$~$\rightarrow$~POS+ADP$_{j}$}
    \ENDIF
\ELSIF{Associated content word is incorrect}
    \STATE $\mathcal{A} =$ \texttt{R:SPELL}
\ENDIF

\IF{Word boundary (WB) error occurs}
    \IF{Words \textit{w$_{i}$w$_{j}$} in source should be separate in target}
        \STATE $\mathcal{A} =$ \texttt{M:WB}
    \ELSIF{Words \textit{w$_{i}$ $\sqcup$ w$_{j}$} in source should be merged in target}
        \STATE $\mathcal{A} =$ \texttt{U:WB}
    \ENDIF
\ENDIF

\IF{Word order error occurs (\textit{w$_{i}$ $\sqcup$ w$_{j}$} in source vs. \textit{w$_{j}$ $\sqcup$ w$_{i}$} in target)}
    \STATE $\mathcal{A} =$ \texttt{R:ORDER}
\ENDIF
\end{algorithmic}
}
\end{algorithm}

Figure~\ref{korean-m2} presents annotations from two annotators for the sentence “비행기 음식이 안 막였습니다” (‘The airplane food didn’t ...’). The noun phrase 음식이 \textit{eumsig-i} (‘food.\textsc{nom}’) is replaced with 음식을 \textit{eumsig-eul} (‘food.\textsc{acc}’), annotated as \texttt{R:NOUN+ADP~$\rightarrow$~NOUN+ADP}, reflecting a case marker substitution from nominative to accusative. The verb 막였습니다 \textit{magyeossseubnida}, a misspelling based on phonetic and orthographic similarity, is corrected by Annotator 0 as 먹었습니다 \textit{meogeossseubnida} (‘ate’) and annotated as \texttt{R:SPELL}. Annotator 1, however, interprets the intended meaning differently and replaces the verb with 맞았습니다 \textit{maj-assseubnida} (‘agreed’), also marked as \texttt{R:SPELL}.

This example highlights a central challenge in grammatical error annotation: distinguishing true grammatical errors from alternative, semantically divergent but grammatically valid expressions. While both corrections are linguistically plausible, they lead to different interpretations of the sentence—either \textit{I didn’t eat the airplane food} or \textit{The airplane food didn’t agree with me}—underscoring the importance of consistency and clarity in annotation guidelines.

\begin{figure}[!ht]
\centering
\footnotesize{
\begin{tabular}{|l|} \hline 
~\\
\texttt{S $_{0}$비행기 $_{1}$음식이 $_{2}$안 $_{3}$막였습니다 $_{4}$.}\\
\texttt{A 1 2|||R:NOUN+ADP~$\rightarrow$~NOUN+ADP|||음식을|||REQUIRED|||-NONE-|||0}\\
\texttt{A 3 4|||R:SPELL|||먹었습니다|||REQUIRED|||-NONE-|||0}\\
\texttt{A 3 4|||R:SPELL|||맞았습니다|||REQUIRED|||-NONE-|||1}\\ 
~ \\
\hline 
\end{tabular}
}

\caption{Examples from the Korean \texttt{M2} file: Annotator 0 corrected the sentence as \textit{I didn’t eat the airplane food}, while Annotator 1 corrected it as \textit{The airplane food didn’t agree with me}, resulting in divergent annotation outcomes. For illustrative purposes, index numbers are included to refer to word positions in the source sentence. These indices are not present in the original \texttt{M2} files but are added here to facilitate discussion, comparison, and reference.}
\label{korean-m2}
\end{figure}

These annotation strategies and enhancements make the KoLLA corpus a robust and linguistically informed resource for Korean GEC. By addressing the structural nuances of the language and capturing correction variability, this enriched dataset offers a valuable foundation for developing and evaluating GEC systems tailored to Korean L2 writing.

\subsection{Rubric-based scoring dataset} 

The relationship between first language (L1) and L2 writing has been extensively explored by L2 writing researchers \citep{hirose-2003-comparing,keck-2006-use,vanweijen-rijlaarsdam-vandenbergh-2019-source}. Recent studies have found that although there are differences in L1 and L2 writing in terms of organizational patterns, overall quality, and grammatical errors, there are also significant similarities between L1 and L2 writing \citep{hirose-2001-persuasive}. 
For example, \citet{fathman-kobayashi-1984-rhetorical} discovered that many Japanese learners of English used a specific-to-general organizational pattern (inductive), which contrasts with American L1 students' general-to-specific writing styles (deductive). However, advanced Japanese ESL students tended to favor the deductive organizational pattern, indicating that L2 writing experience or a certain proficiency level leads to the adoption of organizational patterns preferred in L1 English writing. Similarly, Hirose (2003) compared L1 (Japanese) and L2 (English) organizational patterns in the argumentative writing of Japanese EFL student-writers. The study revealed that a majority of students employed deductive-type organizational patterns in both L1 and L2, challenging the stereotype that Japanese writers naturally favor inductive structures. This finding suggests that with increased proficiency and exposure to L2 writing conventions, learners may adopt organizational patterns aligned with those of native speakers.​Other studies have maintained that not only the quality of writing but also the process of writing might be similar between L1 and L2 writing \citep{davoodifard-2022-overview}, thus justifying the extensive use of L1 process-oriented models of writing in investigating L2 writing \citep{hayes-2000-new,hayes-2012-modeling}. 

Following these findings, this study employs a rubric traditionally used to assess Korean L1 speakers’ writing samples for evaluating Korean L2 learners’ writing. The rubric includes three overarching categories—expression, structure, and content—each with multiple subcategories that provide clear, operationalized criteria for assessment. This rubric is widely used in Korean language education settings and was originally developed and refined by experts in L1 writing assessment to ensure clarity, objectivity, and comprehensiveness in evaluating student performance.

The first category, expression, includes three subcategories: grammatical accuracy, appropriateness of vocabulary use, and appropriateness of expressions. Grammatical accuracy evaluates the correctness of sentence structures, ensuring that sentences are grammatically correct. Appropriateness of vocabulary use measures whether the vocabulary used is suitable for the given context and whether there is a variety of vocabulary used throughout the text. Appropriateness of expressions assesses the variety and suitability of expressions and whether the length of sentences is appropriate.

The second category, structure, consists of four subcategories: appropriateness of internal structure, {organization of essay structure}, coherency of paragraphs, and appropriateness of length. Appropriateness of internal structure measures how cohesively the thesis and supporting sentences are connected. Appropriateness of intra-structure assesses whether the essay includes an introduction, body, and conclusion paragraphs, and whether the length of each paragraph is appropriate. Coherency of paragraphs evaluates the use of conjunctions and transition sentences to ensure smooth flow between ideas. Appropriateness of length checks if the essay meets the required word limit.

The third category, content, includes clarity of topic, detailed explanation, and creativity of thought. Clarity of topic measures whether the topic is clearly presented in the essay and whether each sentence relates to the topic. Detailed explanation assesses whether the explanation is detailed and varied. Creativity of thought examines the uniqueness, logic, and presentation of new and different viewpoints.

Table~\ref{rubric} shows the full rubric used for scoring, including the three main categories—expression, structure, and content—and their respective subcategories and descriptions.

\begin{table}[!ht]
\centering
\resizebox{.98\textwidth}{!}{
\tiny{
\begin{tabular}  {lll} \toprule
{Category} & {Subcategories}  &  {Descriptions} \\ \midrule
{Expression} & \textit{Grammatical Accuracy} & Correct use of sentence structures.\\
& \textit{Appropriateness of Vocabulary Use} & Vocabulary suitability and variety.\\
&  \textit{Appropriateness of Expressions} &  Variety and naturalness of expressions; sentence length appropriateness.\\ \midrule

{Structure} & \textit{Appropriateness of Internal Structure} &  Coherence between thesis and supporting sentences. \\
&  \textit{Organization of Essay Structure} &  Presence and proportionality of introduction, body, and conclusion. \\
&  \textit{Coherency of Paragraphs} &  Logical flow and use of discourse markers.\\
&  \textit{Appropriateness of Length} &  Adherence to required word count. \\ \midrule

{Content} &  \textit{Clarity of Topic} &  Clear presentation of the topic and relevance of content.\\
% &  \textit{Appropriateness of Evidence} (for argumentative essays): Relevance and diversity of supporting evidence.\\
&  \textit{Detailed Explanation} &  Depth and variety in explanation.\\
&  \textit{Creativity of Thought} &  Originality, logical structure, and new perspectives. \\
% &  \textit{Ability to Understand the Prompt} &  Alignment with the given writing prompt. \\
\bottomrule
\end{tabular}
}
}
\caption{Rubric categories and subcategories used for evaluating Korean L2 essays.} \label{rubric}
\end{table}

To ensure that the application of this rubric constituted a standardized assessment procedure, we implemented a multi-step annotator training and calibration process. Two annotators with M.A. degrees in Korean Language Education and extensive teaching and rating experience independently assessed student essays using the rubric. Prior to rating, they underwent two training sessions led by the researcher. In the first session, the rubric categories were thoroughly explained and five sample essays from each proficiency level were collaboratively scored. Discrepancies were discussed in depth until consensus was reached, thereby aligning interpretations of the rubric criteria. In the second session, annotators independently scored essays and reconvened with the researcher to resolve any inconsistencies in interpretation through discussion and clarification.

This process mirrors best practices in standardized assessment, particularly in language testing, where inter-annotator reliability, operationalized rubrics, and calibration sessions are critical for maintaining scoring consistency across annotators and contexts. After independent grading, inter-annotator agreement was measured using Cohen’s kappa.

Table~\ref{kappa} presents the results of Cohen’s kappa, a statistical measure of inter-annotator agreement. This measure accounts for agreement occurring by chance and is commonly used in annotation tasks to evaluate the consistency and reliability of multiple annotators. In our study, we compute Cohen’s kappa across four learner groups: foreign beginner (FB), foreign intermediate (FI), heritage beginner (HB), and heritage intermediate (HI), along with the overall agreement across all groups. The results show strong agreement across all groups, with the highest agreement observed for foreign beginners (0.8253) and the lowest for heritage intermediate learners (0.7245), likely reflecting the relative complexity and fluency of sentences in this group. Overall, the aggregated kappa score of 0.8241 indicates high inter-annotator consistency in our error annotation framework.

Taken together, the use of a detailed, pre-established rubric, combined with structured annotator training and verification of inter-annotator reliability, meets key criteria for a standardized assessment procedure. This provides a strong foundation for using the resulting dataset in further research on L2 Korean writing development.

\begin{table}[!ht]
\centering
{    
\begin{tabular}{c c c c  c}\toprule
FB & FI & HB & HI & All  \\ \midrule
0.8253 & 
0.7921 & 
0.7971 & 
0.7245 & 
0.8241 \\ \bottomrule
\end{tabular}}
\caption{Cohen’s kappa scores across learner subgroups: foreign beginners (FB), foreign intermediates (FI), heritage beginners (HB), and heritage intermediates (HI), as well as the overall agreement (All).} \label{kappa}
\end{table}

In summary, the adoption of a detailed rubric and a carefully implemented annotation protocol ensures a high level of scoring reliability and transparency. These procedures establish the rubric-scored KoLLA corpus as a valuable benchmark for evaluating Korean L2 writing proficiency and developing automated essay scoring systems grounded in pedagogically meaningful criteria.

\section{Findings and Discussion}

The augmented KoLLA corpus reveals valuable insights into both learner error patterns and the pedagogical potential of rubric-based writing evaluation. 
This section discusses these findings in two parts: (i) the implications of multi-reference GEC in capturing linguistic variability in Korean, and (ii) the significance of rubric-based scoring for evaluating overall writing quality and learner proficiency.

\subsection{Multi-reference grammatical error annotation}

The enhancements to the KoLLA corpus provide insights into grammatical error patterns in Korean L2 writing, with particular emphasis on the benefits of multi-reference GEC. To the best of our knowledge, the multi-reference KoLLA corpus represents the first resource of this kind specifically tailored for Korean L2. It addresses inherent linguistic variability in learner writing, thereby enabling more comprehensive and contextually appropriate evaluations.

Single-reference GEC datasets may overlook natural variability in learner language, potentially penalizing alternative yet valid corrections. In contrast, multi-reference annotations capture the diversity of valid grammatical structures, which is particularly critical given Korean's syntactic and morphological flexibility. Korean's syntactic flexibility frequently allows multiple grammatical structures or synonyms to express equivalent meanings without compromising grammatical correctness. Consequently, corrections for case markers, verb endings, and compound verbs often exhibit considerable variability, each reflecting a valid linguistic choice dependent on discourse context.

Lexical flexibility also contributes notably to correction diversity, with many corrections allowing synonymous substitutions without altering intended meaning. This characteristic underscores the importance of flexible evaluation frameworks in GEC systems, prioritizing meaning preservation rather than strict grammatical conformity.

In subsequent analyses, we plan to further investigate specific instances of word-boundary inconsistencies and lexical alternatives by conducting a quantitative error analysis, examining correlations with proficiency levels and evaluating how these variations affect automated correction model performance.

\subsection{Rubric-based scoring for writing assessment}

Beyond grammatical error correction, the introduction of rubric-based scoring provides a broader assessment framework for evaluating Korean L2 writing proficiency. Unlike GEC, which focuses primarily on grammatical accuracy at the sentence level, rubric-based scoring assesses higher-level attributes of writing quality, including coherence, structure, and content development. By incorporating rubric scores aligned with the Korean National Language Institute guidelines, the KoLLA corpus establishes a standardized framework for evaluating L2 writing performance.

Analysis of the rubric scores indicates a strong correlation between grammatical accuracy and overall writing proficiency. Essays with fewer grammatical errors consistently received higher rubric scores, suggesting that error-free writing contributes to improved coherence and clarity. This trend underscores the interdependence of grammatical competence and discourse-level writing skills. Additionally, lexical diversity emerges as a distinguishing factor among proficiency levels. Essays that exhibit a wider range of vocabulary usage tend to receive higher scores, reinforcing the role of lexical variation in language proficiency assessment.

Structural complexity is another key indicator of writing ability, as higher-scoring essays demonstrate stronger paragraph organization, better use of discourse markers, and smoother transitions between ideas. However, despite improvements in grammar and structure at higher proficiency levels, rubric scores reveal that certain discourse-level errors persist even among intermediate learners. Challenges in logical sequencing, underdeveloped arguments, and inappropriate rhetorical structures suggest that targeted instructional interventions are needed to address discourse-level deficiencies in L2 writing.

The rubric-based scoring also highlights differences in proficiency levels across various essay types. Argumentative essays tend to exhibit more discourse errors, particularly in structuring logical arguments and supporting claims with appropriate evidence. In contrast, personal essays often score higher in fluency and coherence but may suffer from lexical repetition. These distinctions indicate that different types of writing require tailored instructional approaches to address specific linguistic challenges.

\subsection{Implications and Future Directions}

The integration of multi-reference GEC and rubric-based scoring in the KoLLA corpus has significant implications for both Korean L2 learning and the development of automated NLP systems. The dataset provides detailed insights into common learner errors, allowing language educators to design targeted curricula that address frequent grammatical challenges such as postposition omission, honorific mismatches, and discourse structure issues. The refined annotation framework offers a valuable resource for the development of instructional materials aimed at improving grammatical competence and writing fluency.

For GEC model development, the inclusion of multi-reference annotations reduces penalization for alternative but grammatically acceptable corrections, making it more suitable for training flexible and context-aware correction models for Korean. By incorporating diverse correction strategies, these models can better accommodate the inherent linguistic variability present in learner writing. Moreover, the multi-reference dataset provides a benchmark for evaluating future GEC systems, ensuring that corrections account for multiple valid grammatical structures rather than rigidly enforcing a single reference.

In the context of AES, rubric-based scores serve as a foundational dataset for training AI-driven assessment models that align closely with human grading standards. By integrating grammatical, lexical, and discourse-level features, these models can provide more nuanced feedback to learners, assisting in both formative and summative assessments of writing proficiency. The incorporation of rubric-based assessments into computational models presents an opportunity to develop more sophisticated evaluation systems capable of providing personalized and structured feedback.

Future research should explore the scalability of multi-reference GEC datasets by incorporating additional annotators to capture a wider range of linguistic variations. Expanding the corpus with discourse-level error annotations could further enhance its utility for both GEC and AES research. Additionally, the integration of rubric scoring into neural network-based assessment models offers promising directions for improving automated feedback systems in Korean L2 writing instruction.

Overall, the findings from this study emphasize the necessity of incorporating both multi-reference grammatical error detection and rubric-based scoring in Korean learner corpora. These enhancements provide a more comprehensive framework for analyzing L2 writing, bridging the gap between error correction, writing assessment, and instructional design.

\subsection{Summary of findings}

The enhanced KoLLA corpus demonstrates the value of integrating both multi-reference grammatical error annotations and rubric-based scoring for a more holistic evaluation of Korean L2 writing. Multi-reference GEC captures the linguistic variability and ambiguity intrinsic to Korean, while rubric-based scoring offers a structured and pedagogically grounded method for assessing broader aspects of writing proficiency. Together, these resources support both computational development and instructional practice, paving the way for more accurate, fair, and informative evaluation of learner language.

\section{Conclusion} \label{conclusion}
This study enhances the KoLLA Korean learner corpus by incorporating multi-reference GEC annotations and rubric-based scoring, addressing critical gaps in Korean L2 writing research. By transforming KoLLA into a multi-reference GEC corpus, we account for the inherent variability in grammatical correction, allowing for more robust evaluations of error patterns and correction strategies. The refined annotation scheme introduces a more structured and linguistically informed approach to error categorization, improving the consistency and accuracy of error analysis. Additionally, the integration of rubric-based scoring provides a standardized framework for assessing writing proficiency, bridging the gap between holistic and detailed linguistic evaluations.

Our findings highlight key grammatical challenges faced by Korean L2 learners, with frequent errors in postpositions, verb conjugations, and word-boundary segmentation. The rubric-based assessments reveal strong correlations between grammatical accuracy, lexical diversity, and overall writing quality, reinforcing the importance of GEC in L2 proficiency development. These enhancements establish KoLLA as a foundational resource for advancing research in language learning, pedagogy, and computational linguistics. The dataset provides insights for educators seeking to design targeted instructional interventions and for researchers developing more effective GEC and automated writing assessment systems.

Several promising directions emerge from this work. Further research into AI-driven rubric scoring systems may refine automated assessment methodologies, enhancing the capacity to provide personalized feedback to learners. Expanding the multi-reference GEC framework to other languages with complex morphological structures could yield broader cross-linguistic insights into second-language acquisition. Moreover, integrating this enriched dataset into AI-powered learning platforms has the potential to facilitate adaptive language instruction, making automated feedback more accessible and effective for Korean L2 learners.

By addressing both linguistic and computational challenges, this study contributes to the ongoing development of Korean learner corpora, reinforcing the importance of high-quality, systematically annotated data in language learning and assessment. The enriched KoLLA corpus serves as a benchmark for future research, paving the way for more sophisticated tools that support Korean L2 learners in achieving greater writing proficiency.

\section*{Data Availability Statement}
The dataset is available at \url{https://doi.org/10.5281/zenodo.15287129} with a DOI and is released under the GNU General Public License Version 3, consistent with the licensing of the original dataset.
% Creative Commons Attribution-NonCommercial-ShareAlike 4.0 International (CC BY-NC-SA 4.0) license.
% For the purpose of anonymous submission and peer review, we have uploaded the proposed dataset as part of the “Related files” in this submission. All identifying information has been anonymized.

\section*{Acknowledgments}
This research  was supported by the 2024 Korean Studies Grant Program of the Academy of Korean Studies (AKS-2024-R-094), awarded to Jungyeul Park.

% \bibliography{references-export}% common bib file

\appendix

\section{Dataset Description}

The enhanced KoLLA dataset comprises two primary components tailored to the distinct goals of grammatical error correction (GEC) and automated essay scoring (AES):

\begin{enumerate}
    \item \textbf{GEC \texttt{M2} File}: This file follows the standardized \texttt{M2} format commonly used in grammatical error annotation tasks. It contains pairwise alignments between original learner sentences and their corrected forms, annotated with detailed error tags based on the extended MRU typology adapted for Korean. Each sentence is associated with two reference corrections, allowing for multi-reference evaluation. Annotations encode the location, type, and nature of each error, capturing both surface-level and structural changes. 

    \item \textbf{AES \texttt{.csv} Files for FB, FI, HB, and HI}: These files contain rubric-based scoring data for four learner groups—foreign beginners (FB), foreign intermediates (FI), heritage beginners (HB), and heritage intermediates (HI). Each CSV file includes 250 columns representing 10 rubric dimensions (Table~\ref{rubric}) across 25 essays. The scoring framework is based on the Korean National Language Institute’s writing evaluation criteria, encompassing expression, structure, and content categories. Each file contains two rows corresponding to independent ratings by two trained annotators. This format allows for inter-rater reliability analysis and enables fine-grained evaluation of writing performance across demographic and proficiency subgroups.
\end{enumerate}

\end{document}